# Hybrid methodology for hourly global radiation forecasting in Mediterranean area

Cyril Voyant[1,2*], Marc Muselli[1], Christophe Paoli[1], Marie-Laure Nivet[1]

[1] University of Corsica, CNRS UMR SPE 6134, 20250 Corte, France

[2] Castelluccio Hospital, Radiotherapy Unit, BP 85, 20177 Ajaccio, France

*Corresponding author. Address: Université de Corse, UMR SPE 6134, Route des Sanguinaires, 20000 AJACCIO, France. Tel.: +33 4 95 52 41 30; fax: +33 4 95 45 33 28. E-mail address: cyrilvoyant@gmail.com (C. Voyant).

**Abstract.**

The renewable energies prediction and particularly global radiation forecasting is a challenge studied by a growing number of research teams. This paper proposes an original technique to model the insolation time series based on combining Artificial Neural Network (ANN) and Auto-Regressive and Moving Average (ARMA) model. While ANN by its non-linear nature is effective to predict cloudy days, ARMA techniques are more dedicated to sunny days without cloud occurrences. Thus, three hybrids models are suggested: the first proposes simply to use ARMA for 6 months in spring and summer and to use an optimized ANN for the other part of the year; the second model is equivalent to the first but with a seasonal learning; the last model depends on the error occurred the previous hour. These models were used to forecast the hourly global radiation for five places in Mediterranean area. The forecasting performance was compared among several models: the 3 above mentioned models, the best ANN and ARMA for each location. In the best configuration, the coupling of ANN and ARMA allows an improvement of more than 1%, with a maximum in autumn (3.4%) and a minimum in winter (0.9%) where ANN alone is the best.

**Nomenclature**

| Symbol | Description | Symbol | Description |
|---|---|---|---|
| $CI(t)$ | Clearness index at time $t$ | $CSI(t)$ | Clear sky index at time $t$ |
| $X(t)$ | Radiation time series at time $t$ [Wh/m²] | $CSI\text{-}PC(t)$ | Clear sky index corrected by periodic coefficients at time t |
| $H_0(t)$ | Extraterrestrial solar radiation coefficient [Wh/m²] | $p,q$ | Order of ARMA(p,q) |
| $I_{SC}$ | Solar constant [Wh/m²] | $L$ | Lag operator |
| $E_0$ | Eccentricity of the Earth trajectory | $\varphi_t, \theta_t$ | $\varphi_i$, Parameters of the AR model, $\theta_i$, Parameters of the MA model |
| $\delta$ | Declination [rad] | $\rho_i$ | Autocorrelation factor estimation for time lag i |
| $\phi$ | Latitude [rad] | $\rho_{ii}$ | Estimation of partial autocorrelations for time lag $i$ |
| $\omega_i$ | Hour angle [rad] | $R$ | Cross-correlation estimation |
| $H_{gh,clearsky}$ | Clear sky global horizontal radiation [Wh/m²] | $endo^e$ | $e$ lags endogenous for ANN |
| $\tau$ | Global total atmospheric optical depth | $N^n$ | $n$ lags of nebulosity for ANN |
| $b$ | Fitting parameter of the Solis clear sky model | $P^{ps}$ | $ps$ lags of pressure for ANN |
| $h$ | Solar elevation angle (rad) | $RP^{pc}$ | $pc$ lags of precipitation for ANN |
| $nbH$ | Number of effective hour uses in one year (h) | $\hat{x}^{AR/ANN}(t)$ | Prediction with AR and ANN model of the global radiation at time (t) [Wh/m²] |
| $\hat{x}^{AR}_{spr/sum}(t)$ | Prediction with AR model of the global radiation at time (t) and with a training set limited to the spring and summer seasons [Wh/m²] | $\hat{x}^{ANN}_{aut/win}(t)$ | Prediction with ANN model of the global radiation at time (t) and with a training set limited to the autumn and winter seasons [Wh/m²] |
| $IC(t)$ | Interval Confidence at time $t$ | | |

# 1  Introduction

Solar radiation is one of the principal energy sources for physical, biological and chemical processes, occupying the most important role in many engineering applications. The process of converting sunlight to electricity without combustion allows to create power without pollution. It is certainly one of the most interesting topics in solar energy area. Thereby it is necessary to create some prediction models to use ideally this technology. By considering their effectiveness, it will be possible for example to identify the most optimal locations for developing a solar power project or to maintain the grid stability in solar and conventional power management system. Thus the solar energy forecasting is a process used to predict the amount of solar energy available in the current and near terms. The horizon we studied in this article is directly related to the renewable PV energy integration. Indeed, the intermittence of this renewable energy source penalizes and limits its incorporation on the grid. In France for example, the threshold is fixed to 30% related to the total electrical energy supplied. In order to modify this boundary, authorities and managers need tools allowing to estimate the energy deficit (associated to cloud occurrence concerning the PV part of the total renewable energy). There are essentially two solutions; the first is related to the storage of the previous excess energy in order to smooth the PV production and the second is related to a rapid switch (done by the grid manager) to a non-intermittent energy source (often from fossil fuel in PV case). It is essential to anticipate the global radiation decrease (or increase) for an ideal transition. The time delay between power up and electricity provision depends on the technique used. For example in the case of gas turbine, the running time is 30 minutes and about 1 hour for combustion engines. So it is necessary for the grid manager to know the PV energy prediction at least one hour ahead. Several methods have been developed by experts around the world and the mathematical formalism of Times Series (TS) has been often used. TS is a set of numbers that measures the status of some activity over time. It is the historical record of some activity, with measurements taken at equally spaced intervals with a consistency in the activity and the method of measurement. Some of the best predictors found in literature are Auto-Regressive and Moving Average (ARMA) [1-2], Bayesian inferences [3-4], Markov chains [5-6], k-Nearest-Neighbors predictors [7-8] or artificial intelligence techniques as the Artificial Neural Network (ANN) [9-11]. Although these methodologies are potentially good in many areas, we observed in our previous studies on global radiation prediction [12-13] that ANN and ARMA methods were the most interesting. We have also demonstrated that an optimized ANN with endogenous and exogenous inputs can forecast the global solar radiation time series with acceptable errors. However ARMA was very competitive and even outperformed ANN in some cases. So we decided to investigate the contribution of a hybrid model to the global radiation prediction problem, combing ANN and ARMA. Many authors have already studied successfully the coupling between ANN and different traditional computing technologies in solar energy area. Most of the time, thee technologies used are fuzzy logic, wavelet-based analysis, genetic algorithm, ARMA and ANN methods [10,14-17]. The main reason for this success seems to be the synergy of each computational property that makes them suitable for particular problems and not for others. In our case, we have coupled a linear process (ARMA) which describes well the "ordinary" solar radiation (clear sky) and a non-linear process with endogenous and exogenous input (ANN). Thus, we developed a hybrid ANN-ARMA method which seems more appropriate to predict the cloudy skies ("extraordinary sequence"). Inspired by biological neural networks the ANN methodology is able to solve a variety of problems in decision making, optimization, control and obviously prediction and more particularly time series prediction [9, 18-21]. As the univariate ARMA model, ANNs can be considered as one of the referenced predictors among t ime series forecasting methods.

The paper is organized as follow: Section 2 describes the data we have used. In section 3, before proposing a hybrid model of ARMA and ANN techniques, we recall the need to make stationary a time series and we present three approaches to make stationary data. Finally to conclude this section we present preliminary results comparing ANN and ARMA, which led us to imagine a hybrid method. The section 4 includes final results and experiences conducted during this study and showing that forecasting results can significantly be improved by selecting ANN or ARMA models according to their performances. The section 5 concludes and suggests perspectives.

## 2   The data

The study proposes a new method to forecast solar radiation in Mediterranean area for the next hour (h+1 prediction). The approach is based on the hypothesis that global solar radiation can be computed using endogenous and exogenous meteorological parameters. The interest of this horizon is primordial in the case of conventional power management to maintain the grid stability (in France and more specifically in Corsica 25% of the electrical consumption is related to renewable energy).

The radiation time series (Wh.m$^{-2}$) are measured for five seaside meteorological stations maintained by the French meteorological organization (Météo-France) which offersaccess to data: Ajaccio (41°5'N and 8°5'E, 4 m.a.s.l), Bastia (42°3'N, 9°3'E, 10 m.a.s.l), Montpellier (43.6°N and 3.9°E, 2 m.a.s.l), Marseille (43.4°N and 5.2°E, 5 m.a.s.l) and Nice (43.6°N and 7.2°E, 2 m.a.s.l). These stations are equipped with pyranometers and standard meteorological sensors (pressure, cloudiness, etc.). The stations are located near the sea with relief nearby and under a Mediterranean climate characterized by hot summers with abundant sunshine and mild, dry and clear winters. This specific geographical configuration makes cloudiness difficult to forecast. The data representing the global horizontal solar radiation were measured on an hourly basis from January 1998 to December 2007 (10 years). The first eight years have been used to parameterize our models and the last two years to test them. Note that in the presented study, only the hours between 8:00AM and 04:00PM (true solar time) are considered, the others hours are not interesting from the energy point of view, and it is very complicated to make stationary the measures of sunrise and sunset with a ratio to trend (multiplicative scheme detailed in the next section). Note that, although only the data for nine daytime hours are considered the time series used during all the experiments are continue without gaps for the rejected samples, e.g. the data pair for t=9 (last hour of the first day) and t=10 (first hour of the second day) is handled as any other data pair through the day. A first treatment allows us to clean the series of non-typical points related to sensor maintenances or absence of measurement. Less than 4 % of measurements were missing and replaced by the hourly average for the given hour. The other meteorological parameters studied are pressure ($P$, Pa; hourly average and gradient[*]), cloudiness ($N$, Octas) and rain precipitations ($RP$, mm). Although other weather data were available, we focused on these three variables that have been most significant in the experiments carried out in previous works [22-23]. Figure 1 shows the localization in Mediterranean area of the five weather stations studied.

---

[*] Difference between the mean pressure of hour $h$ and hour $h-1$

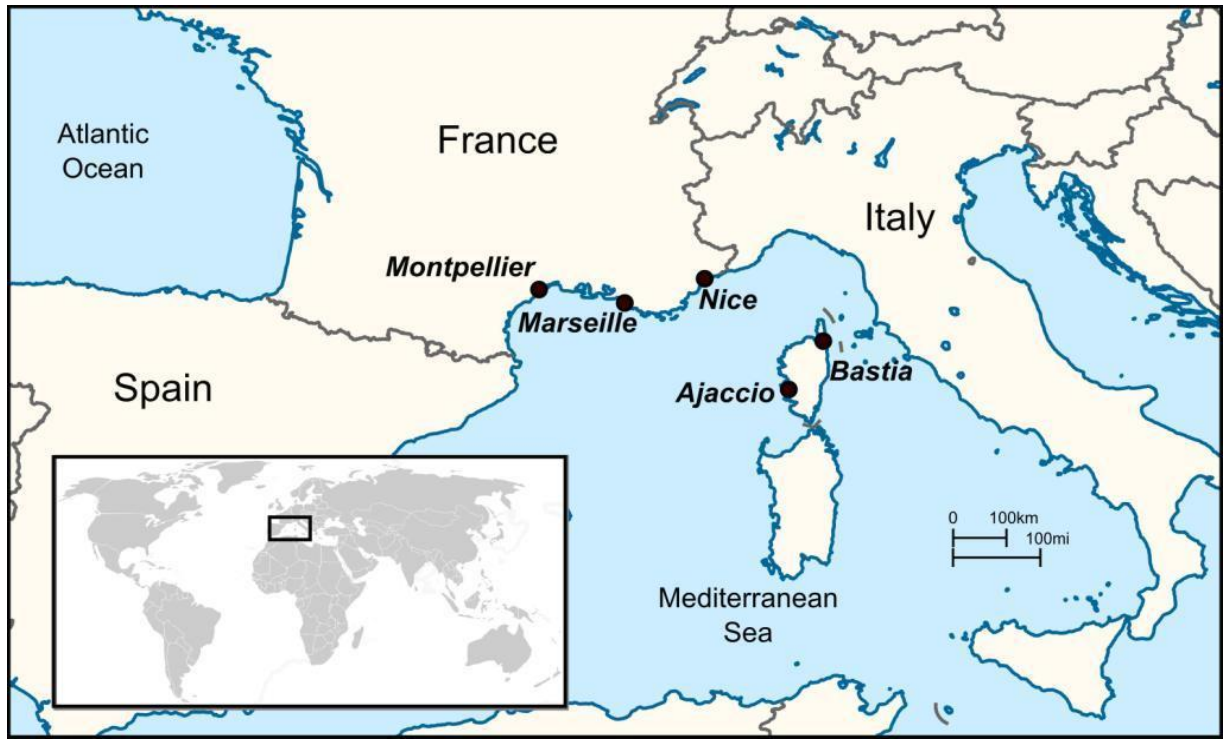

5   Figure 1. The five studied stations marked in the Mediterranean sea: Ajaccio, Bastia , Montpellier , Marseille  and Nice.

## 6   Time series forecasting models

A time series [21] is a collection of time ordered observations $x_t$, each one being recorded at a specific time t (period). Time series can appear in a wide set of domains such as Finance, Production or Control, just to name a few. In first approximation, a time series model ($\hat{x}_t$) assumes that past patterns will occur in the future. In fact a time series model could be used only to provide synthetic time series statistically similar to the original one. The modeling of the series begins with the selection of a suitable mathematical model (or class of models) for the data. Then, it is possible to predict future values of measurements [24]. Before proposing a hybrid ANN/ARMA model, we recall the need to make stationary a time series. We present three approaches to make stationary data available (the comparison of these methodologies will be presented in section 4 dedicated to the results). Then we recall the definitions and principles of ARMA and ANN techniques and their generic configuration which led us to build a hybrid method presented in the last sub-section.

## 6.1 The need to stationarize

The prediction of the solar energy time series on the earth's surface can be perturbed by the non-stationarity of the signal and the periodicity of the phenomena. While ARMA methods explicitly require that the data are stationary, there are very few references in the literature and sometimes contradictory [25-27] relating the benefit of making data stationary in ANN methodologies. In our case, we have used physical phenomena in an attempt to overcome the seasonality of the resource (determinist component). We propose three approaches to make stationary data available: the first is based on the Clearness Index (CI) [28], the second on a Clear Sky Index (CSI) [29], the third is a variant of the previous one with Periodic Coefficients (PC) [2].

The first methodology based on the Clearness Index (CI(t)) is defined as the ratio between the total ground radiation (called X(t)) and extraterrestrial radiation ($H_0(t)$) :

$$CI(t) = X(t)/H_0(t) = X(t)/\left(I_{sc}E_0\left(\sin\delta(t)\sin\phi + \cos\delta(t)\cos\phi\cos\omega_i(t)\right)\right) \qquad \textbf{Equation 1}$$

The required parameters to compute CI(t) are classical incelestial mechanics [2]. The second methodology is based on the clear sky model with the CSI. Several methods allow to determine this model. In our case, we have used the simplified "Solis clear sky" [30] model based on radiative transfer calculations and the Lambert-Beer relation [31]. This expression of the atmospheric transmittance is valid with polychromatic radiations, however when dealing with global radiation, the Lambert-Beer relation is only an approximation because of the back scattered effects. According to Mueller et al. [30], this model remains a good fitting function of the global horizontal radiation. The use of this model requires fitting parameter (b), extraterrestrial radiation ($H_0(t)$) and solar elevation (h). In this case, the clear sky global horizontal irradiance ($H_{gh,clearsky}$) reaching the ground is defined by:

$$H_{gh,clearsky}(t) = H_0(t).e^{-(\tau/\sin^b(h(t)))}.\sin(h(t)) \qquad \textbf{Equation 2}$$

We have validated the Solis model on a horizontal global radiation with a series of tests considering one year of daily solar radiation data that are not presented in this paper [10]. We obtain a relation of stationarization where X(t) is the global radiation measure and CSI(t) is the new time series (clear sky index):

$$CSI(t) = X(t)\Big/H_0(t).e^{-(\tau/\sin^b(h(t)))}.\sin(h(t)) \qquad \textbf{Equation 3}$$

The third methodology is also based on CSI but adding Periodic Coefficients (CSI+PC). Indeed, experiences show that the CSI can be not very convincing for some hours of some days. This is especially true, when the shading mask changes measurements, the CSI is not effective, and do not allow to make the time series stationary. Like in economic research studies [2], it is possible to use the Periodic Coefficients (PC) in order to overcome the stationary problem like, for example, this daily single effects related to the shading mask. The method consists to estimate trend by moving averages and fixed coefficients. The chronological series is composed of n cycles (10 years), each comprising p periods (9x365 hours = 3285 hours = nbH in the following, the other hours of the day are taken equal to zero). Firstly, we proceed by eliminations of seasonality following a multiplicative scheme:

a) Estimation of trend by calculating centered moving averages over p periods (MM(t) is the average);
b) Calculation of CSI'(t) = CSI(t) / MM(t) ratio;
c) Calculation of the seasonal adjustment (from periodic coefficients; PC) of previous ratio by average on
   PC(t) = < CSI'(t)>$_{10\ years}$ (the PC(t) is a series of nbH numbers);
d) Calculation the TS seasonally corrected by the ratio: CSI *(t) = CSI(t) / PC(t) (modulo nbH).

Thus, the new series (CSI *(t)) is without seasonality, due to these seasonal coefficients. The impact of the use of seasonal adjustments is apparent on the autocorrelogram (Figure 2).

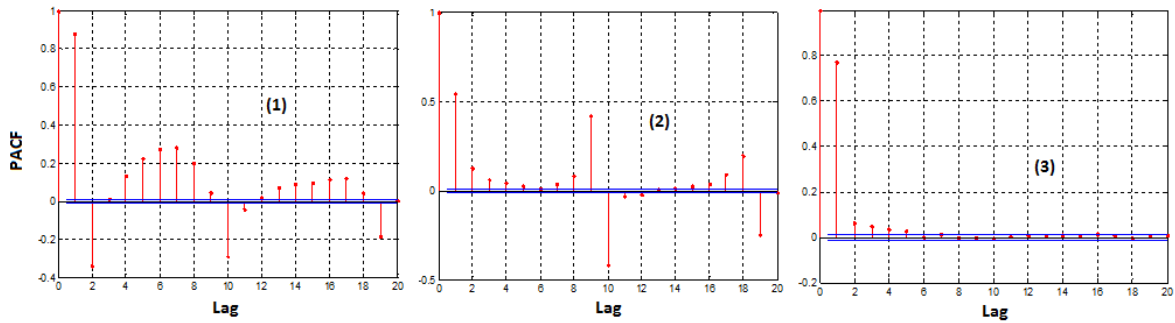

Figure 2. Autocorrelogram of the time series studied for the Ajaccio station. 1-original series ;2-CSI ;3-CSI and PC.

We can observe the effect of stationarization on the time series. On the original series, we can see that there is a daily periodic component, the use of ratio to trend (CSI) decreases the links between the hours CSI(t) and CSI(t-1), ...., CSI(t-8), that represent the stationarity within the day, but the daily seasonality is not corrected (peak at t-9 and t-10). Only with the use of periodic coefficients (CSI+PC), we get a series made stationary. However, Figure 2 does not mean that the CSI+PC method is the best for our methodology and the other two methods have to be eliminated, more particularly in the case of ANN. Indeed, all three methods will involve different settings on ANNs and therefore different architectures. That is why the impact of the three methods on the quality of prediction is presented in section dedicated to the results.

## 6.2 The ARMA model

The ARMA techniques [1] are reference estimators in the prediction of global radiation field. It is a stochastic process coupling an autoregressive component (AR) to a moving average component (MA). The model is usually then referred to as the generic ARMA(p,q) model where p is the order of the autoregressive part and q is the order of the moving average part. Finally, the combined ARMA (p,q) model is given by the following equation (with the lag operator L formalism, e.g. $L^i x_t = x_{t-i}$):

$$(1 - \sum_{i=1}^{p} \varphi_i L^i).x_t = (1 + \sum_{i=1}^{q} \theta_i L^i).\varepsilon_t \qquad \text{Equation 4}$$

Where, $x_t$ is a time series, $\varphi$ and $\theta$ are the parameters of the autoregressive and moving average parts, L is the lag operator and ε is an error term distributed as a Gaussian white noise. The optimization of these parameters must be made depending on the type of the series studied. In this study, we chose to use Matlab© software. The criterion adopted to consider when an ARMA model "fits" to the time series is the normalized Root Mean Square Error (nRMSE) described by the equation 5 (x represents the measurement and y the prediction):

$$nRMSE = \sqrt{<(x-y)^2>/(<(x)^2>)} \qquad \text{Equation 5}$$

This measure is generated by the prediction of two years of radiation not used during the ARMA parameters calculation step (years 2006-2007). Residual autocorrelogram tests have been computed to verify white noise error terms. Table 1 shows for each of the five stations the optimal ARMA type obtained, the values of $\varphi_1$ and $\varphi_2$ and some features as the value of the loss function and the Akaike's Final Prediction Error (FPE). These both last features are often used with ARMA modeling, the first is related to the quadratic error and the second simulates the cross validation situation where the model is tested on another data set (this notion outperforms the classical nRMSE or loss function because the overfitting and the model complexity are take into account). The algorithm used is the Yule-Walker fitting method [1].

| Station | ARMA type (p,q) | $\varphi_1$ | $\varphi_2$ | Loss function | Akaike's FPE |
|---|---|---|---|---|---|
| Ajaccio | (1,0) | 0.5435 | 0 | 0.078 | 0.078 |
| Bastia | (1,0) | 0.5838 | 0 | 0.077 | 0.077 |
| Montpellier | (1,0) | 0.5178 | 0 | 0.099 | 0.099 |
| Marseille | (2,0) | 0.4176 | 0.1350 | 0.090 | 0.090 |
| Nice | (1,0) | 0.5248 | 0 | 0.087 | 0.087 |

Table 1. Optimal ARMA and some features obtained for each of the five stations with the centered CSI.

We can observe that we obtained a different ARMA(p,q) type for Marseille. However all the models presented are very simples: less than 2 parameters (p≤2 and q=0) are sufficient models. Moreover, we can see that values of the loss function and the Akaike's FPE are equal, obviously because the ARMA model are constructed with few parameters and the prediction test covers a wide period. Note that, these both features are only done for information and their values were not used during the optimization step. The results of nRMSE are done in the section 4. Moreover, Table 1 shows only the result for CSI, concerning the CSI+PC and non processed series (original series) the results are equivalent.

### 6.3 ANN for time series prediction

Artificial neural networks (ANN) are intelligent systems that have the capacity to learn, memorize and create relationships among data [9]. An ANN is made up by simple processing units, the neurons, which are connected in a network by a large number of weighted links where the acquired knowledge is stored and over which signals or information can pass. We have chosen to study a particular ANN architecture called Multi Layer Perceptron (MLP) because it has been the most used architecture both in the renewable energy domain and in the time series forecasting [25, 32-34]. A well known, difficult and consuming-time task is to find the best network configuration and its parameters. The more common and important accepted optimization parameters are: number of input neurons, number of hidden layer and neurons, type of activation functions, type of normalization of data and the type of learning algorithm. In recent studies [12, 13, 22, 23], we have developed an optimization process composed by three independent and chronological subparts:

1. Choice of the endogenous lags (time-delays between network nodes) number: past values of the global radiation time series are added as inputs of the ANN architecture;
2. Choice of the exogenous lags numbers for each meteorological parameter available: cloudiness, pressure, precipitation;
3. Optimization of the ANN: data normalization, hidden neurons number, parameters of the learning algorithm, etc.

The
**Figure** 3 presents the generic MLP architecture obtained after several experiments cited above. The obtained characteristics are: one hidden layer, the activation functions are hyperbolic tangent (hidden) and linear (output), the Levenberg-Marquardt learning algorithm (with max fail parameter equal to 5 to the validation, µ decrease

and increase respectively to 0.1 and 0.001, and goal equal to zero). Inputs are normalized on {-1,1}. The Matlab© training, validation and testing data sets were respectively set to 90%, 10% and 0% for the first eight years (1998-2005). The test has been done separately on the last two years (2006-2007). So the ratio is 72% for training, 8% for validation and 20% for testing.

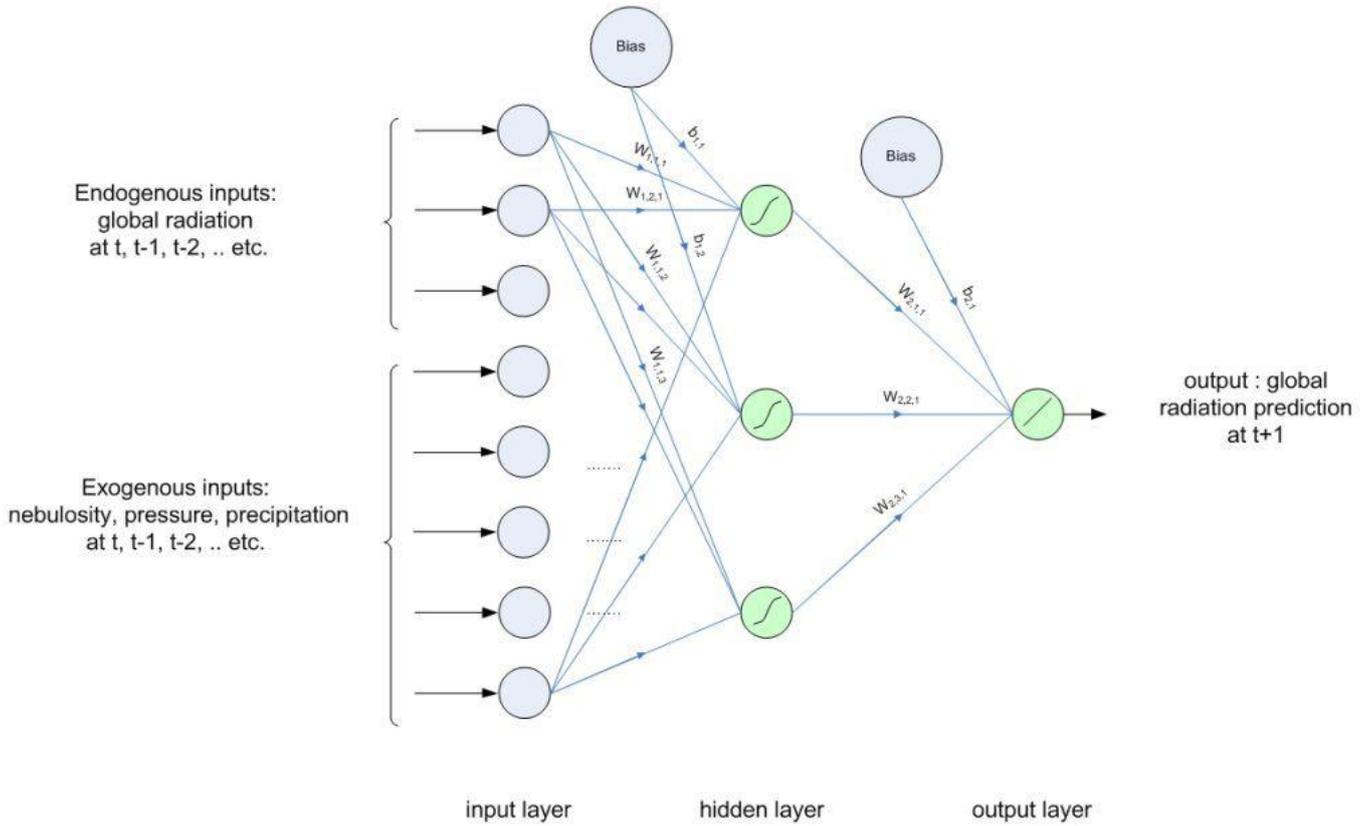

Figure 3: Generic architecture of the MLP

From this generic architecture, it is now possible to start our optimization process. Thus the first step of this process consists in determining the number of endogenous time lags. We have used autocorrelation criteria and a validation test based on the student T-test [35]. Only the lags with an autocorrelation different from zero are considered and put as input of the network. The partial autocorrelations (PACF noted $\rho_{ii}$), are obtained from theoretical autocorrelation factor (ACF $\rho_i$) equation. Generally it is easier to use its recursive formula:

$$\rho_{ii} = \begin{cases} \rho_1 & \text{if } i = 1 \\ \dfrac{\rho_i - \sum_{j=1}^{i-1} \rho_{i-1}\rho_{i-j}}{1 - \sum_{1}^{i-1}\rho_{i-1}\rho_j} & i = 2...k \end{cases}$$ **Equation 6**

The second step of the optimization process consists in determining the number of exogenous input nodes (exogenous data are not pretreated with a stationarization process). For this purpose, the correlation measure is computed in order to determine which of the exogenous parameters are to consider. The correlation between two variables (cross-correlation) reflects the degree to which the variables are linked (considering the limitation that correlation criteria can only detect linear dependencies between variables). The most common correlation measure is the Pearson's correlation. A correlation of +1 (or -1) means that there is a perfect positive (or negative) linear relationship between variables and a value of 0 implies that there is no linear correlation between the variables. The Pearson correlation coefficient (R) between two variables is defined from covariance and variance of the two variables. For a series, the estimation of R is given by:

$$R = \sum_{k=1}^{N}(x_k - \bar{x})(y_k - \bar{y}) \bigg/ \sqrt{\sum_{k=1}^{N}(x_k - \bar{x})^2 \sum_{k=1}^{N}(y_k - \bar{y})^2}$$ **Equation 7**

Generally, a Pearson correlation between -0.5 and 0.5 indicates a weak or a non association between 2 variables. The link to the Student test (T-test) shows that the score *R* may be used as a statistic test to assess the significance of a variable. In our experiments, the limit of significance according to the T-test indicates a threshold very low. Indeed, the limit is below 0.1 for the sample above 1000 elements and for an alpha level of 0.05. Obviously, this methodology is not appropriate in our case because the threshold for the coefficient *R* should be more important to select a limited number of exogenous inputs. Indeed we have chosen an *R* threshold equal to 50% for the cloudiness and 15% for other. Thereby only the higher correlation will be chosen. In the cloudiness case, a threshold below 50% is chosen because it is the average of the PACF in the first lag. The cloudiness and the global radiation are similar: when one increases, the other decreases. So to be considered, the exogenous data must provide additional information not already presents in the endogenous data. A threshold below 50% suggests that no additional information is available. Concerning exogenous parameters P and RP, adjustable thresholds have been tested. We have noted that a threshold more than 20% is so restrictive that no exogenous data would be eligible. On the other hand, the decrease of this threshold (less than 10%) is responsible of the increase of the nodes number, making the net too complicated. Taking into account both the parsimony and the efficiency for the ANN, a threshold equal to 15% represents the best compromise.

The last and third step of the optimization process consists in determine the number of neurons in the hidden layer. For this, we used a method based on the experience: the configuration giving the best results is kept for the following experiments. An example of the result of the overall optimization process is given in the Figure 4: ten hidden neurons were chosen in this case and the 95% Confidence Interval (CI95%) was computed after eight simulations.

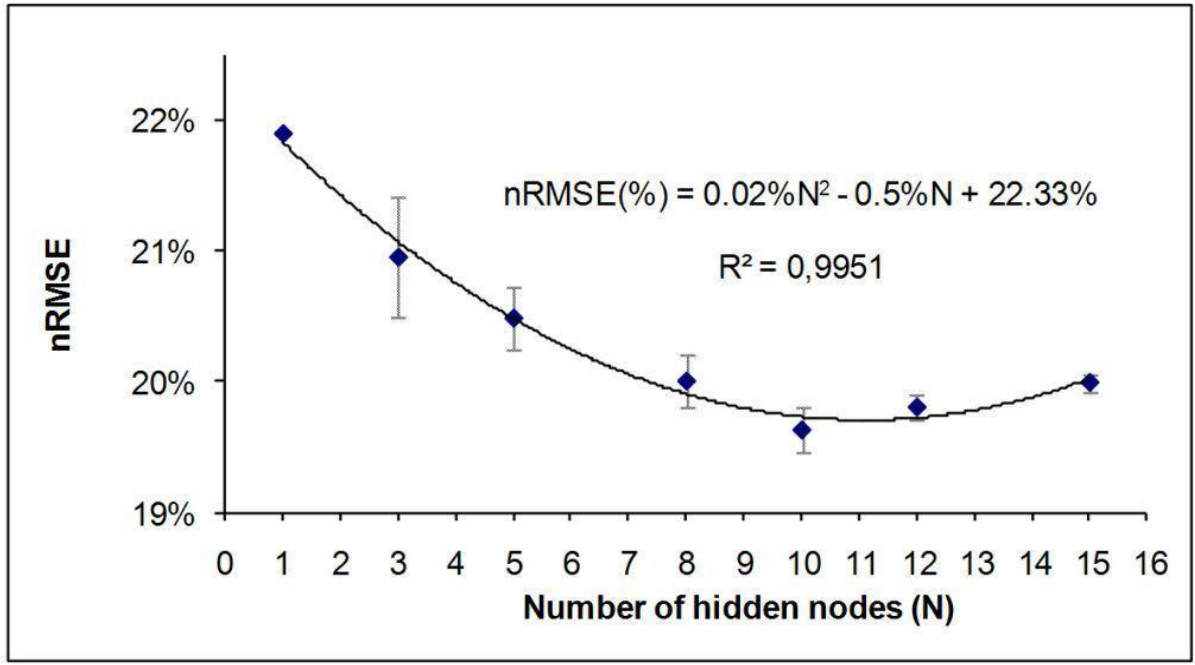

Figure 4. Example of optimization for the number of neurons for only one hidden layer.

For example, in the case of the Ajaccio station, we obtained a MLP with 10 endogenous inputs (radiation at time t, t-1, t-2, …, t-9), 2 exogenous inputs (cloudiness at time t and t-1), 1 hidden layer with 15 neurons and 1 output neuron (irradiation at time t+1). In the following we adopt a canonical form for the ANN architecture: $(endo^e, N^n, P^{ps}, RP^{pc}) \times H \times S$, where *e, n, ps* and *pc* are the number of neurons in relation with endogenous data, cloudiness, pressure and rain precipitation. H and S are respectively the neuron number of the hidden layer and the output layer. Table 2 presents the canonical forms of the best exogenous and endogenous ANN architectures obtained for each of the five stations studied.

| Station | Best exogenous architecture | Best endogenous architecture |
|---|---|---|
| Ajaccio | $(endo^{10}, N^2, P^1, RP^1) \times 15 \times 1$ | $(endo^{10}) \times 15 \times 1$ |
| Bastia | $(endo^{10}, N^2, P^1, RP^1) \times 15 \times 1$ | $(endo^{10}) \times 10 \times 1$ |
| Montpellier | $(endo^{10}, N^2, P^1) \times 10 \times 1$ | $(endo^{10}) \times 10 \times 1$ |
| Marseille | (endo$^{10}$,N$^1$, P$^1$)x15x1 | (endo$^{10}$)x10x1 |
| Nice | (endo$^{10}$,N$^1$, RP$^1$)x15x1 | (endo$^{10}$)x10x1 |

Table 2. Best ANN exogenous and endogenous architectures obtained for each of the five stations and with CSI methodology.

It should be noted that the best results were obtained with the CSI methodology: all the details of the comparison of stationarization methodologies are presented in the Table 3 of the section 4 dedicated to the results.

### 6.4    Building a hybrid method with ANN and ARMA

Based on the results summarized in previous sub-sections and in our previous studies, we propose to combine Artificial Neural Network (ANN) and Auto-Regressive and Moving Average (ARMA) models from simple rules based on the analysis of hourly data series. Three models are proposed: the first proposes simply to use AR models for six months in spring and summer and to use an optimized ANN for the other part of the year (model A); the second model is equivalent to the first but with a seasonal learning (model B); the last model depends on the error occurred the previous hour (model C). Figure 5 shows the overall scheme of the proposed hybrid methods.

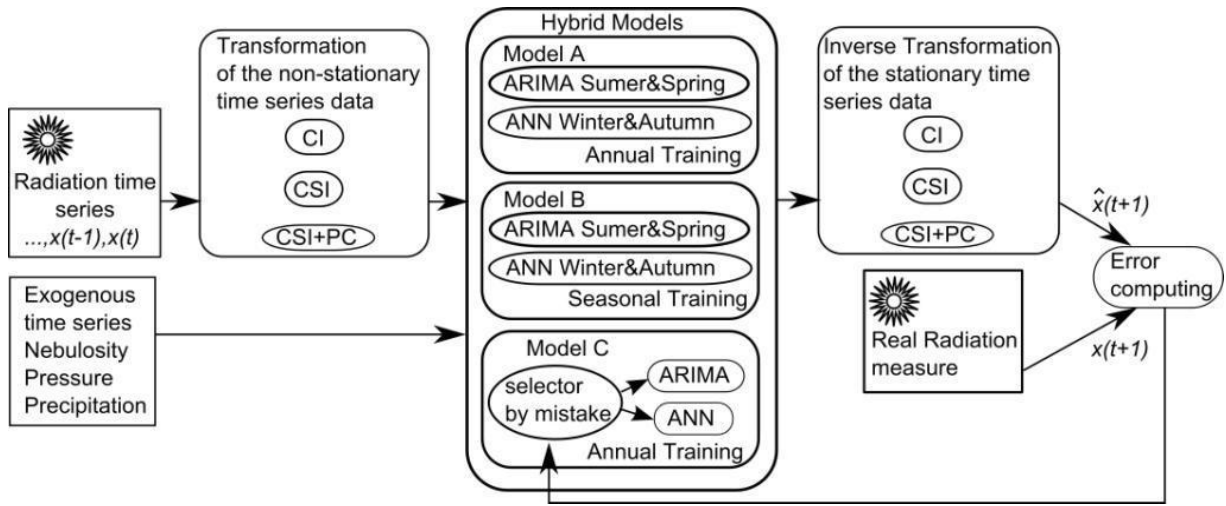

Figure 5. Overall scheme of the proposed hybrid methods.

The first model studied, A, is deterministic: six months of the year are predicted with an AR and the other six months with an ANN with endogenous and exogenous inputs. We can summarize this method with the following rule:

$$if\ t \in \{spring\ OR\ summer\}\ then\ \hat{x}(t+1) = \hat{x}^{AR}(t+1)\ else\ \hat{x}(t+1) = \hat{x}^{ANN}(t+1) \qquad \text{Equation 8}$$

The second model, B, is equivalent to the first, but learning is seasonal: the learning of ANN is done only in autumn and winter and for the AR only in the spring and summer. We can summarize this method with the following rule:

$$if\ t \in \{spring\ OR\ summer\}\ then\ \hat{x}(t+1) = \hat{x}^{AR}_{spr/sum}(t+1)\ else\ \hat{x}(t+1) = \hat{x}^{ANN}_{aut/win}(t+1) \qquad \text{Equation 9}$$

The third model, denoted Model C, is a stochastic model which depends on the prediction error at the previous hour. We can summarize this method with the following rule where ε is the residue of the prediction:

$$if\ |\varepsilon^{AR}(t)| \leq |\varepsilon^{ANN}(t)|\ then\ \hat{x}(t+1) = \hat{x}^{AR}(t+1)\ else\ \hat{x}(t+1) = \hat{x}^{ANN}(t+1) \qquad \textbf{Equation 10}$$

All these rules have been elaborated by the observations of the results obtained with the mono-predictors ANN and ARMA. Furthermore the next section presents results obtained using the three hybrid models on the five stations studied.

## 7 Experiments and results

This section includes all the results and experiments conducted during this study. First, we present the influence of the three stationarization methodologies on ANN and ARMA configurations. Then, we compare performance of ANNs and ARMA models against a persistence model (e.g. persistence of irradiance; the irradiance at time t-1 is equal to irradiance at time t), a clear sky model and a simple average through a basic method of ranking. The result of this ranking has formed the basis of the hybrid method presented in the previous section. Finally, we present results obtained with hybrid models on the 5 stations studied.

### 7.1 Influence of the time series stationarity on ANN configuration

In this sub-section, we present and compare the influence of the three methodologies presented in section 3.1 on an ANN predictor. As ARMA methods require explicitly that data have to be stationary and many works have already treated this problem [1-2], we do not present the influence of the stationarization methodologies on ARMA methods. However, it should be noted that the best results for ARMA models were obtained with the CSI methodology but results obtained with CSI or CSI+PC were quite similar.

On the Table 3, the performances of the three stationarization methodologies are compared against an ANN trained with a non-stationary time series: CV is the coefficient of variation (dispersion rate; ratio between the standard deviation and the mean) and the error is represented by the nRMSE (defined from expectation value E by $nRMSE = \sqrt{E[(X-\hat{X})^2]}/\sqrt{E[X^2]}$ ; best nRMSE are in bold). We can see that the CSI methodology minimizes the prediction error for the five cities studied, followed by the CI methodology, the CSI+PC methodology and the non-stationary time series. In fact, contrary to an ARMA predictor, stationarity does not affect much the prediction error (gap ~ 0.5% of nRMSE). In addition, the interpretation of CV is difficult: we see that if its value is high, the prediction error is high, but the opposite is not observed on this table.

| Station Lat/Long/Alt | Stationarity | ANN Architecture | CV | nRMSE |
|---|---|---|---|---|
| Ajaccio 41.5°/8.5°/4m | none | $(endo^{10},N^2)$x15x1 | 0.61 | 0.171 |
| | CI | $(endo^{10},N^3,P^1)$x15x1 | 0.44 | 0.169 |
| | CSI | $(endo^{10},N^2,P^1,RP^1)$x15x1 | 0.46 | **0.167** |
| | CSI + PC | $(endo^1,N^2,P^1,RP^1)$x5x1 | 0.38 | 0.172 |
| Bastia 42.3°/9.3°/10m | none | $(endo^{10},N^2)$x10x1 | 0.64 | 0.201 |
| | CI | $(endo^{10},N^3)$x15x1 | 0.46 | 0.199 |
| | CSI | $(endo^{10},N^2)$x10x1 | 0.48 | **0.196** |
| | CSI + PC | $(endo^1,N^2)$x3x1 | 0.44 | 0.200 |
| Montpellier 43.6°/3.9°/2m | none | $(endo^{10},N^1)$x15x1 | 0.63 | 0.170 |
| | CI | $(endo^{10},N^3)$x15x1 | 0.45 | 0.165 |
| | CSI | $(endo^{10},N^2,P^1)$x10x1 | 0.49 | **0.164** |
| | CSI + PC | $(endo^1,N^2,P^1)$x5x1 | 0.40 | 0.168 |
| Marseille 43.4°/5.2°/5m | none | $(endo^{10},N^1)$x10x1 | 0.60 | 0.155 |
| | CI | $(endo^{10},N^2,P^1)$x15x1 | 0.42 | **0.149** |
| | CSI | $(endo^{10},N^1, P^1)$x15x1 | 0.45 | **0.149** |
| | CSI + PC | $(endo^1,N^1,P^1)$x8x1 | 0.35 | 0.151 |
| Nice 43.6°/7.2°/2m | none | $(endo^{10}, N^1)$x15x1 | 0.63 | 0.169 |
| | CI | $(endo^{10},N^2)$x10x1 | 0.44 | 0.169 |
| | CSI | $(endo^{10},N^1, RP^1)$x15x1 | 0.46 | **0.166** |
| | CSI + PC | $(endo^1,N^1, RP^1)$x3x1 | 0.38 | 0.172 |

Table 3. Comparison of CI, CSI and CSI + PC on the forecasting against a non-stationary time series.

Furthermore we can see that CSI+PC methodology makes neural networks architecture much simpler: only a single neuron endogenous and a maximum of 8 neurons in the hidden layer. However the prediction errors are not improved and it seems that there is no need to improve the CSI methodology with PC in this context of prediction. In the ARMA methodology the results obtained with CSI and CSI+PC are also similar. In the next only the CSI methodology has been used for ANN or ARMA.

### 7.2 Comparison of forecasting models ANN and ARMA

In this sub-section, we compare the performances of ANNs, ARMA, persistence and clear sky models and a simple average. The interested reader can find more details on these models in [22]. Concerning the ANN models, we have studied a MLP with only endogenous variables and a MLP with endogenous and exogenous variables. In ARMA, we used the Yule Walker algorithm derived from the analysis of correlograms available on the Matlab© software. For ARMA and ANN, the data are made stationary with the use of CSI methodology as proposed in section 4.1.

For each of these models, we chose to consider annual results but also seasonal result in order to have a better comprehension of the phenomena. Thus, it can be seen in Table 4 that there is no interest to use methods like ANN with endogenous and exogenous variables in summer. Indeed, because there are rarely clouds during this period, a linear process like ARMA seems sufficient. We can therefore conclude that the use of ANN with exogenous variables is only interesting during short periods where a lot of clouds on the sky are observed (essentially in autumn and winter). However Bastia, Montpellier and Nice are the city where the annual error is minimized mainly by the use of a simple auto-regression (respectively nRMSE=17.7%, nRMSE=16% and nRMSE=16.2%).

| Station | Model | Annual | Winter | Spring | Summer | Autunm |
|---|---|---|---|---|---|---|
| Ajaccio | Persistence | 35.3 | 55.9 | 32.2 | 32.3 | 36.9 |
| | Clear Sky | 41.8 | 66.9 | 39.4 | 37.9 | 43.2 |
| | Average | 32.5 | 51.3 | 29.9 | 28.6 | 33.9 |
| | ARMA | 16.9 | 25.7 | **15.9** | **14.4** | 19.3 |
| | ANN endo | 17.1 | **23.8** | 16.3 | 15.8 | 18.3 |
| | ANN exo | **16.7** | **23.8** | 16.3 | 15.8 | **17.6** |
| Bastia | Persistence | 39.6 | 50.2 | 41,9 | 34,5 | 37,7 |
| | Clear Sky | 45,9 | 62,4 | 48,7 | 38,9 | 42,5 |
| | Average | 35,7 | 48,1 | 37,3 | 29,8 | 36,5 |
| | ARMA | **17.7** | 23.4 | **18.2** | **15.7** | **18.8** |
| | ANN endo | 19.4 | 22.5 | 21.5 | 17.2 | 19.9 |
| | ANN exo | 19.6 | **22.4** | 21.4 | 17.6 | 20.1 |
| Montpellier | Persistence | 37.3 | 44.1 | 36.0 | 30.5 | 47.2 |
| | Clear Sky | 43.2 | 55.9 | 36.8 | 34.0 | 61.3 |
| | Average | 34.9 | 41.9 | 32.6 | 27.9 | 47.8 |
| | ARMA | **16.0** | 20.1 | **14.4** | **13.3** | 20.5 |
| | ANN endo | 16.5 | 18.6 | 14.6 | 15.0 | 18.3 |
| | ANN exo | 16.4 | **18.3** | 14.7 | 15.1 | **17.8** |
| Marseille | Persistence | 32.7 | 40.1 | 31.9 | 23.5 | 43.7 |
| | Clear Sky | 40.0 | 56.0 | 38.2 | 31.6 | 49.7 |
| | Average | 29.9 | 39.9 | 28.9 | 21.3 | 39.7 |
| | ARMA | 15.4 | 21.4 | 14.8 | **12.5** | 19.7 |
| | ANN endo | 15.2 | **18.3** | 14.5 | 13.3 | 18.1 |
| | ANN exo | **14.9** | **18.3** | **14.4** | 12.8 | **17.0** |
| Nice | Persistence | 35.2 | 41.6 | 39.4 | 29.2 | 41.8 |
| | Clear Sky | 40.8 | 51.5 | 41.7 | 33.2 | 48.3 |
| | Average | 31.7 | 40.2 | 33.6 | 27.1 | 38.4 |
| | ARMA | **16.2** | 20.6 | **16.2** | 14.4 | 18.6 |
| | ANN endo | 16.8 | 18.5 | 18.3 | 16.7 | 15.9 |
| | ANN exo | 16.6 | **18.2** | 17.7 | 16.3 | **15.0** |

Table 4. Performance comparison (nRMSE in % and in bold for the best configurations) of ANNs, ARMA models, persistence model, a clear sky model and a simple average for the hour and the day considered.

We see also, that no predictor seems unanimous but that persistence, clear sky simulation and average methods are the worst predictors (nRMSE close to or greater than 30% for all the studied towns). It is noted that rules to establish what the best predictor is, appears to depend on the season. To identify these rules, we have developed a simple method of ranking in which the best predictor has the smallest nRMSE. Thus, we have assigned a point rating system: 1 point for the best predictor, 2 points for the second, ... and 6 points for the worst. We hypothesized that the prediction error produces the same damage for a network manager regardless of the season. The interest to weight prediction methods according to the seasons will be broached as perspectives in the conclusion section. The operation was repeated for the five stations and for each season (see table 5). The best predictor is the one that has the fewest score: minimum is 5 points and maximum is 30 points.

| Models | Winter | Spring | Summer | Autumn |
|---|---|---|---|---|
| ANN exo | $1^{st}$ (5 pts) | $2^{nd}$ (10 pts) | $2^{nd}$ (12 pts) | $1^{st}$ (7 pts) |
| ANN endo | $2^{nd}$ (8 pts) | $3^{rd}$ (12 pts) | $2^{nd}$ (12 pts) | $2^{nd}$ (10 pts) |
| ARMA | $3^{rd}$ (15 pts) | $1^{st}$ (7 pts) | $1^{st}$ (5 pts) | $3^{rd}$ (13 pts) |
| Average | $4^{th}$ (20 pts) | $4^{th}$ (20 pts) | $4^{th}$ (20 pts) | $4^{th}$ (20 pts) |
| Persistence | $5^{th}$ (25 pts) | $5^{th}$ (25 pts) | $5^{th}$ (25 pts) | $5^{th}$ (25 pts) |
| Clear Sky | $6^{th}$ (30 pts) | $6^{th}$ (30 pts) | $6^{th}$ (30 pts) | $6^{th}$ (30 pts) |

Table 5. Ranking and scores obtained by the predictors depending on the season.

The clear sky model, the persistence and the average are for the 5 city ranked last. During seasons of high cloud cover (winter and autumn), an ANN with exogenous variables (meteorological data) gives the best results. Concerning the two other seasons where clouds are rarer (spring and summer), the ARMA model is the most efficient. These results encouraged us to develop a hybrid model based on the analysis of hourly data series and seasons presented in previous section and whose results are presented below.

### 7.3 Results of the hybrid models

Table 6 shows the results obtained by the 3 hybrid models (A, B, C) on the 5 stations studied. The comparison of performances is computed in % of nRMSE annually and for each season.

| Station | Model | Annual | Winter | Spring | Summer | Autumn |
|---|---|---|---|---|---|---|
| Ajaccio | A | 15.9 | 23.8 | 15.9 | 14.4 | 17.6 |
|  | B | 16.4 | 24.3 | 15.9 | 14.4 | 17.6 |
|  | C 2157 ARMA / 4413 ANN | **15.3** | **21.1** | **15.0** | **14.3** | **16.0** |
| Bastia | A | **17.7** | 23.4 | 18.2 | 15.7 | 18.8 |
|  | B | 17.9 | 22.4 | 18.2 | 15.7 | 20.0 |
|  | C 2675 ARMA / 3895 ANN | **17.7** | **21.3** | **19.0** | **16.1** | **17.9** |
| Montpellier | A | 15.5 | 18.3 | 14.4 | **13.3** | 17.8 |
|  | B | 15.7 | 19.1 | 14.5 | 13.4 | 18.4 |
|  | C 2293 ARMA / 4277 ANN | **15.1** | **17.2** | **13.5** | 13.4 | **17.1** |
| Marseille | A | 14.7 | 18.3 | 14.8 | 12.5 | 17.0 |
|  | B | 15.0 | 19.4 | 14.8 | 12.4 | 18.7 |
|  | C 2000 ARMA / 4570 ANN | **13.7** | **17.5** | **13.2** | **11.5** | **16.0** |
| Nice | A | 15.3 | 18.2 | **16.2** | **14.4** | 15.0 |
|  | B | 15.7 | 19.5 | **16.2** | **14.4** | 16.2 |
|  | C 2077 ARMA / 4493 ANN | **15.1** | **16.2** | 16.3 | 14.9 | **14.2** |

Table 6. Seasonal performance comparison (nRMSE in % and in bold for the best results) of hybrid models.

We can see that the results have improved significantly through the use of hybrid methods A and C. Method B, is certainly as effective, but the truncating data learning penalizes this method. It would certainly try it with a quantity of data more consistent to really test it. Considering annual results, method C is the most effective for the five stations, the distribution of predictions is 1/3 for ARMA and 2/3 for ANN. We see that this method saves more than 1% related to the non-hybrid methods (see table 3). The figure 6 shows the average gain of nRMSE related to the use of the hybrid (model C) compared to the best ANN (dark gray bar) and AR (gray bar) for all the station. The maximum is in autumn (3.4% better than ARMA) and the minimum is in winter, when the hybrid method is less interesting than an ANN method (0.9% worse than ANN). With AR in summer, this is the only case where the hybrid method is less interesting than conventional forecaster.

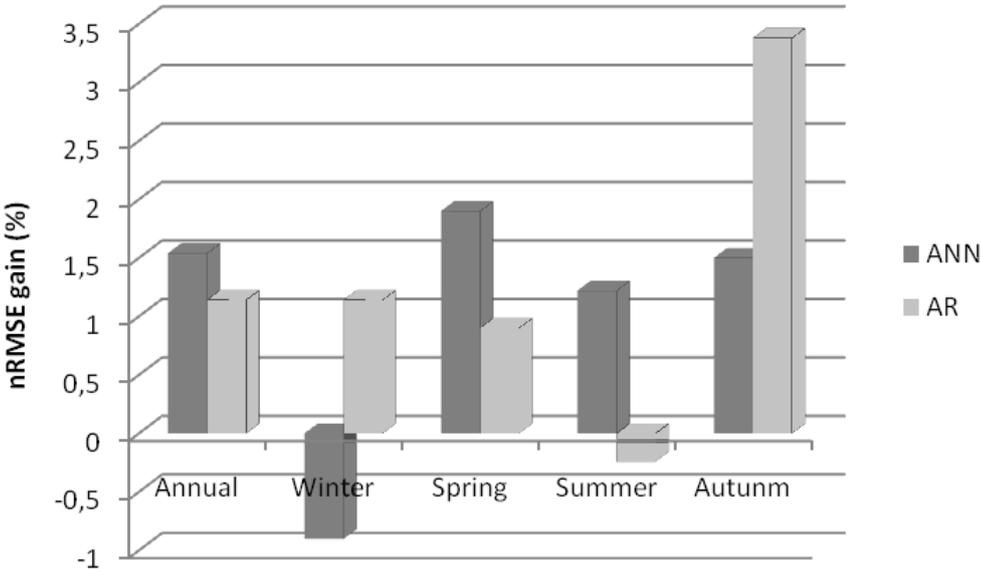

Figure 6. Average gain of nRMSE related to the use of the hybrid model C compared to ANN and AR.

The Figure 7 (on the left) presents the comparison between measures and forecast with Model C for the hourly global horizontal radiation (curve type y=x). We see quite clearly that the model C is very effective to predict global radiation: the coefficient of determination between the simulated data and the measurements is about 0.95.

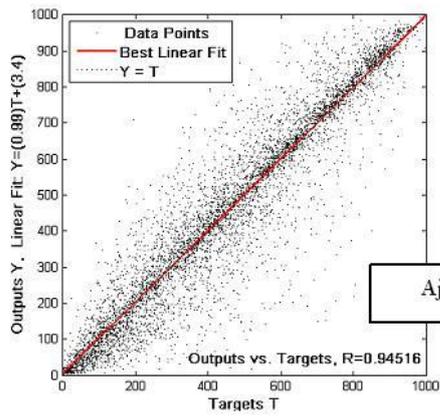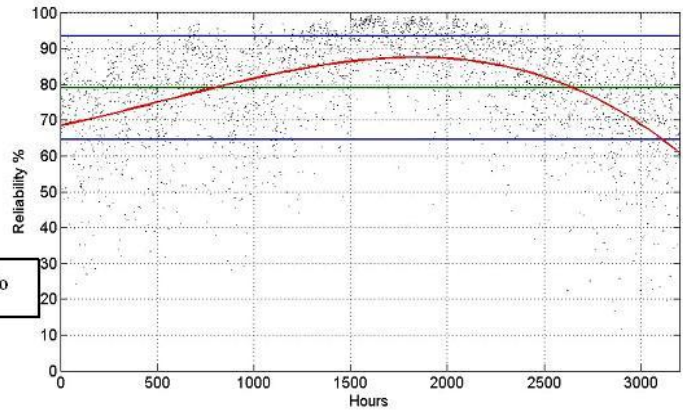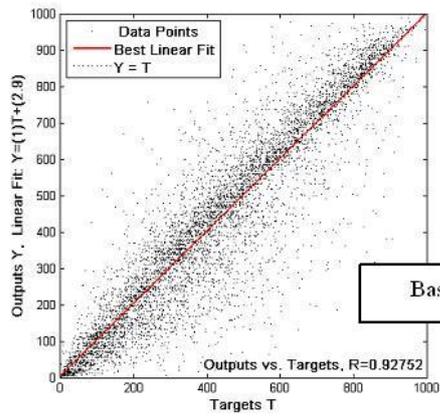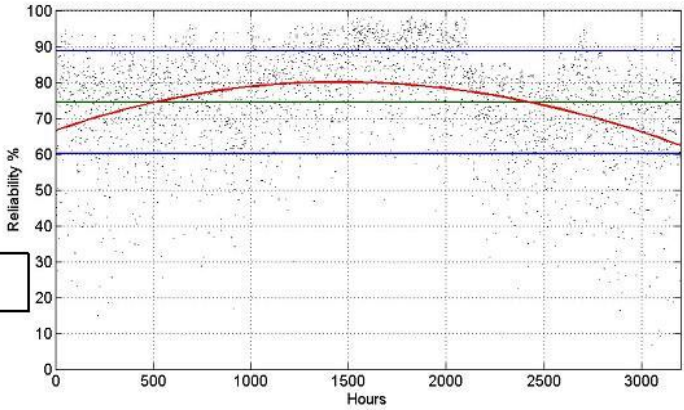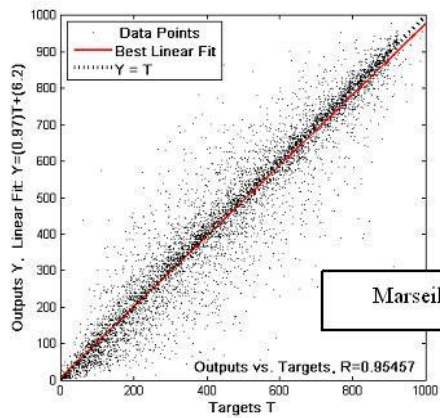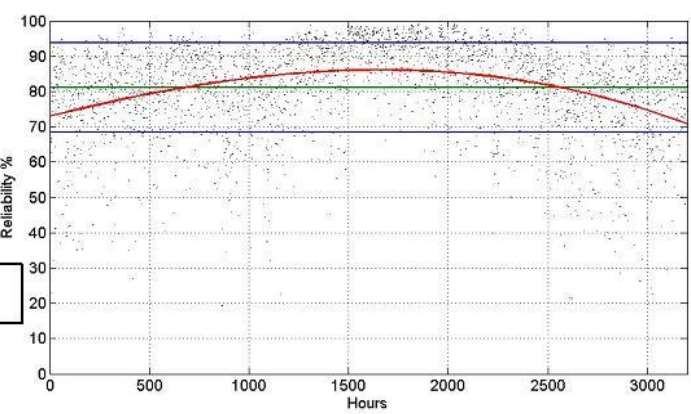

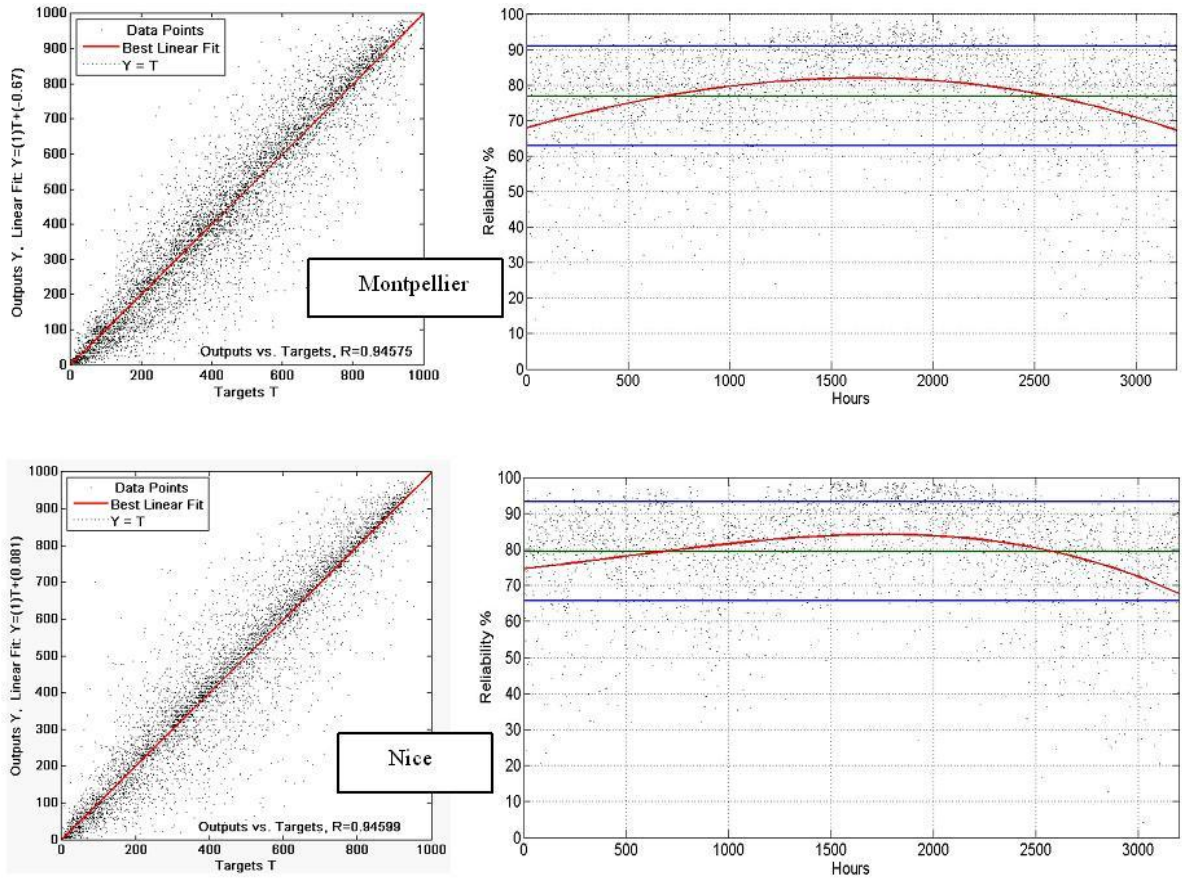

Figure 7. Comparison between measures and forecasts with Model C (left), and reliability of the prediction for the hybrid model C (right, see text for description levels).

To characterize the quality of hybrid model we have developed a confidence parameter. Each prediction calculated from the hybrid model is linked to a reliability index η(t): 100% represents the fact that we are sure of the prediction value, 0% corresponds to an error higher than the real radiation value. This computing is done during the training step of the MLP. During the prediction process, the simulator gives for each hour two parameters: the h+1 horizon global radiation and the confidence we can give to this value. The computing method of the reliability index is represented by the Eq. 9 (the negative values which correspond to a prediction error of 100% are replaced by 0, e.g. if $|\hat{x}(t) - x(t)| > x(t)$, then $\eta(t) = 0$, else $\eta(t) =$ equation 11). An average done on the training sample allows to transform this series to a new series of efficiency, for each hour and each day of the year, of 3285 elements (3285=365x9).

$$\eta(t) = 100.\left[1 - |\hat{x}(t) - x(t)| / x(t)\right] \qquad \textbf{Equation 11}$$

Figure 7 (on the right) presents the reliability of the prediction for the hybrid model C: the 3500 hours on the axis corresponds to the year of prediction, the green line is the average, the red line is the trend and the blue lines are the average +/- standard error. We see on this figure that the maximum one-year trend of reliability is situated in the summer, and it exceeds 80%. In winter, it drops below 70%. The average annual reliability is about 80% but some predictions are accompanied by very low reliability, especially in winter (<30%).

Figure 8 presents also the comparison between the prediction (red crosses) and the measures (black line) for a period between June and July 2006. The confidence interval (average ± IC) is computed from the reliability index according to Eq. 12:

$$IC(t) = \hat{x}(t).\left[-(1/100).\eta(t) + 1\right] \qquad \textbf{Equation 12}$$

This transformation is a renormalization of the reliability: thus, $\eta(t) = 100\%$ corresponds to a $IC(t) = 0$, and $\eta(t) = 0\%$ correspond to a $IC(t) = \hat{x}(t)$. The IC represents the average error computed for each hour of the year during the training step of the MLP. This parameter is so computed through a height years interval that is an important historical sample making the average exploitation possible. The IC used is not ideal, but allows to quantify the hours of the year where usually the prediction is good or on the contrary bad. This information is very important for a grid manager who needs predictions. Note that this parameter (IC) should not be confused with the "conventional" Confidence Interval (called CI) used in statistics and estimated from the standard deviation, the confidence level and the sample size.

The first part of the windows (Figure 8) is typically the profile of sunny days without cloud cover (hours 1 to 37). The model is very efficient. Some hours have very low values of radiation (< 400 Wh/m²) but the forecast with the hybrid methodology is very powerful. Concerning the central and last parts of the plot where the profiles seem indicate cloudy days (hours > 37), we can see that the model is again reliable. Note that in general the prediction ± IC includes correctly the measure but sometimes not at all (hour 43 or 41). This is the consequence of the IC computing methodology which is based on an average prediction error (during MLP training): the extraordinary cloud occurrences (in time or opacity) are not included in the model.

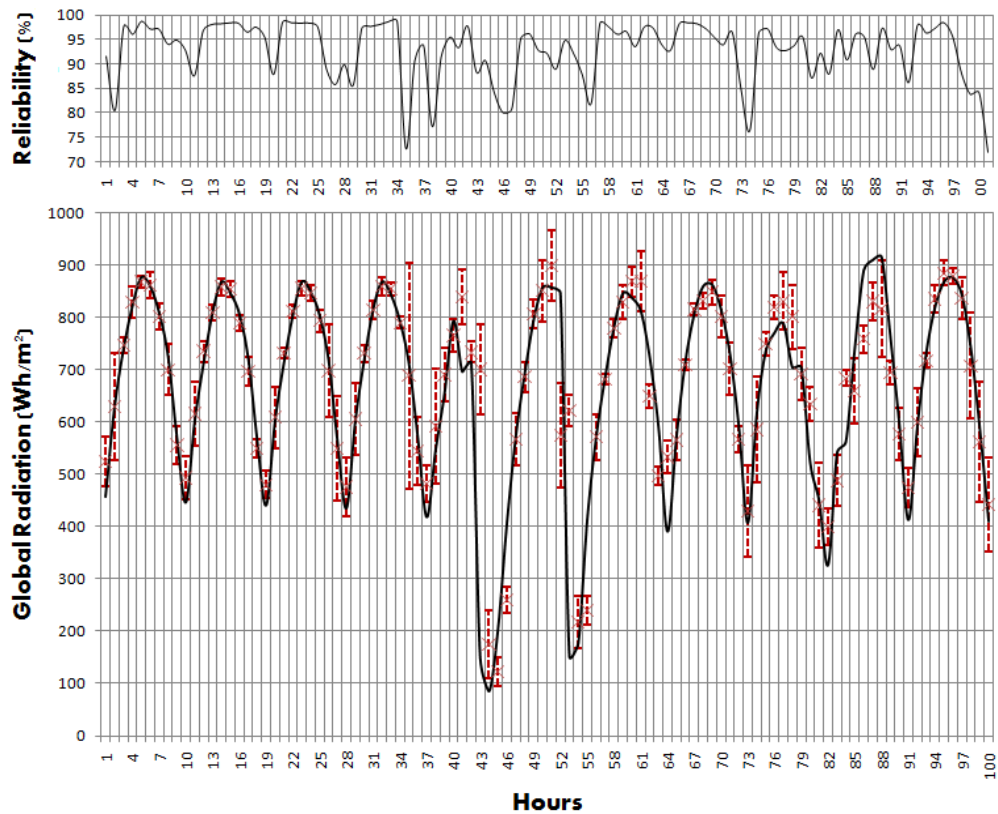

9  Figure 8. Comparison between the prediction (red crosses with confidence intervals) and the measures (black line). The corresponding reliability of each measure is done on the upper curve (period between the 200[th] and the 211[th] day (June-July 2006).

## 10  Conclusion

In this paper, we proposed to combine Artificial Neural Network (ANN) and Auto-Regressive and Moving Average (ARMA) models from simple rules based on the analysis of hourly data series. We used data taken from five meteorological ground stations located in Mediterranean area. Before to find the best ARMA and ANN configurations, three approaches to make stationary the time series have been presented and compared. A methodology based on the clear sky index has been selected because it minimized the prediction error for the five cities studied. From these data made stationary, we developed best ARMA and ANN configurations from which three hybrid models have been proposed and driven: the first model proposes simply to use ARMA for six months in spring and summer and to use an optimized ANN for the other part of the year; the second model is equivalent to the first but with a seasonal learning; the last model depends on the error occurred the previous hour. These models were used to forecast time series data of hourly global horizontal radiation for five places in Mediterranean area. The forecasting performance was compared among several models: the three above mentioned models, the best ANN and ARMA for each location. In the best configuration, the coupling of ANN and ARMA allows an improvement of more than 1%, with a maximum in autumn (3.4%) and a minimum in winter where an ANN alone is the best (0.9%). These results show clearly that the complexity of the extended methodology presented here, does not yet allow a meaningful gain. Ideally, we should ask each grid manager from when the gain is significant (policy energy and installed capacities of available PV farms should for example be taken into account). If this gain is about 3%, the use of hybrid method (like presented here) is certainly irrelevant, the alone ANN method would be more appropriated. On the contrary, for a gain widely upper than 3% the hybrid method would be relevant. All this shows the limits of forecast procedures purely based on onsite dataIn future studies, it would be very relevant to add to our optimized ANN new data as geographical features or forecast meteorological data, derived from the ALADIN model for example. In addition, the study of an approach for forecasting 24-h of solar irradiance using several MLP connected has to be done. Finally, as noted in sub-section 4.2, it should be interesting to study the possible contribution of Multiple Criteria Decision Aiding (MCDA) methods [36] in order to rank several prediction methods. The use of these methods in the solar energy domain is not new [37-38] and they can be a scientific basis for the definition of decision criteria that have to be elaborated with energy managers.

## Acknowledgements

This work was partly supported by the CTC (Collectivité Territoriale de Corse). We thank the French National Meteorological Organization (Météo-France) which has supervised the data collection from their data bank for the five synoptic stations.

**List of captions:**

**List of tables**